\documentclass{ecai} 

\usepackage{siunitx}
\usepackage{latexsym}
\usepackage{amssymb}
\usepackage{amsmath}
\usepackage{amsthm}
\usepackage{booktabs}
\usepackage{enumitem}
\usepackage{graphicx}
\usepackage{color}
\usepackage{adjustbox}
\usepackage{multirow}
\usepackage{diagbox}
\usepackage[table]{xcolor}

\usepackage[justification=raggedright,singlelinecheck=false]{caption}

\newcommand{\BibTeX}{B\kern-.05em{\sc i\kern-.025em b}\kern-.08em\TeX}

\begin{document}


\begin{frontmatter}


\paperid{123} 


\title{AI Fairness Beyond Complete Demographics: Current Achievements and Future Directions}

\author[A]{\fnms{Zhipeng}~\snm{Yin}}
\author[A]{\fnms{Zichong}~\snm{Wang}}
\author[A]{\fnms{Avash}~\snm{Palikhe}}
\author[B]{\fnms{Zhen}~\snm{Liu}}
\author[C]{\fnms{Jun}~\snm{Liu}}
\author[A]{\fnms{Wenbin}~\snm{Zhang}\thanks{Corresponding author. Email: wenbin.zhang@fiu.edu}}

\address[A]{Florida International University, Miami, Florida, United States}
\address[B]{Guangdong University of Foreign Studies, Guangzhou, Guangdong, China}
\address[C]{Carnegie Mellon University, Pittsburgh, Pennsylvania, United States}


\title{AMCR: A Framework for Assessing and Mitigating Copyright Risks in Generative Models}





\begin{abstract}Generative models have achieved impressive results in text to image tasks, significantly advancing visual content creation. However, this progress comes at a cost, as such models rely heavily on large-scale training data and may unintentionally replicate copyrighted elements, creating serious legal and ethical challenges for real-world deployment. To address these concerns, researchers have proposed various strategies to mitigate copyright risks, most of which are prompt based methods that filter or rewrite user inputs to prevent explicit infringement. While effective in handling obvious cases, these approaches often fall short in more subtle situations, where seemingly benign prompts can still lead to infringing outputs. To address these limitations, this paper introduces Assessing and Mitigating Copyright Risks (AMCR), a comprehensive framework which i) builds upon prompt-based strategies by systematically restructuring risky prompts into safe and non-sensitive forms, ii) detects partial infringements through attention-based similarity analysis, and iii) adaptively mitigates risks during generation to reduce copyright violations without compromising image quality. Extensive experiments validate the effectiveness of AMCR in revealing and mitigating latent copyright risks, offering practical insights and benchmarks for the safer deployment of generative models.

\end{abstract}

\end{frontmatter}


\section{Introduction}

Generative models have emerged as a powerful tool for advancing visual content creation, particularly through text-to-image tasks~\citep{borji2022generated,liao2025deepseek,pallant2024creating,rao2024generative,amon2024uncertain}. Their capability to generate high-quality images directly from textual descriptions has facilitated widespread adoption across diverse fields, such as digital marketing, creative design, education, and entertainment~\citep{alhabeeb2024text,anantrasirichai2022artificial,lee2024impact,vartiainen2023using,yin2025Uncertain}. For example, in digital marketing, carefully designed textual prompts allow these models to rapidly produce tailored visual content, significantly reducing production time and costs while simultaneously enhancing creative flexibility. Consequently, generative models have become crucial tools in modern visual content generation processes. 

However, despite their success in generating desired images from user-provided prompts, these models raise significant copyright concerns. Specifically, generated outputs frequently incorporate characteristics from their training datasets~\citep{hartwig2024survey,saharia2022palette,wang2025fdgen}, potentially leading to images that closely resemble original artworks. Such similarities have triggered legal disputes, as demonstrated by lawsuits filed by artists against both Stability AI and MidJourney, accusing them of utilizing artists’ works to train their models without obtaining consent, thereby enabling the replication of distinctive artistic styles and creations~\citep{hayes2023generative}. Likewise, in the United Kingdom, Getty Images, Inc. has filed a lawsuit against Stability AI, alleging the unauthorized use of Getty's proprietary images to train the Stable Diffusion model~\citep{samuelson2023ongoing}. These developments underscore the urgent need for research on responsible text-to-image generation practices that effectively address copyright-related challenges.

\begin{figure}[htbp]
\includegraphics[width=1\linewidth]{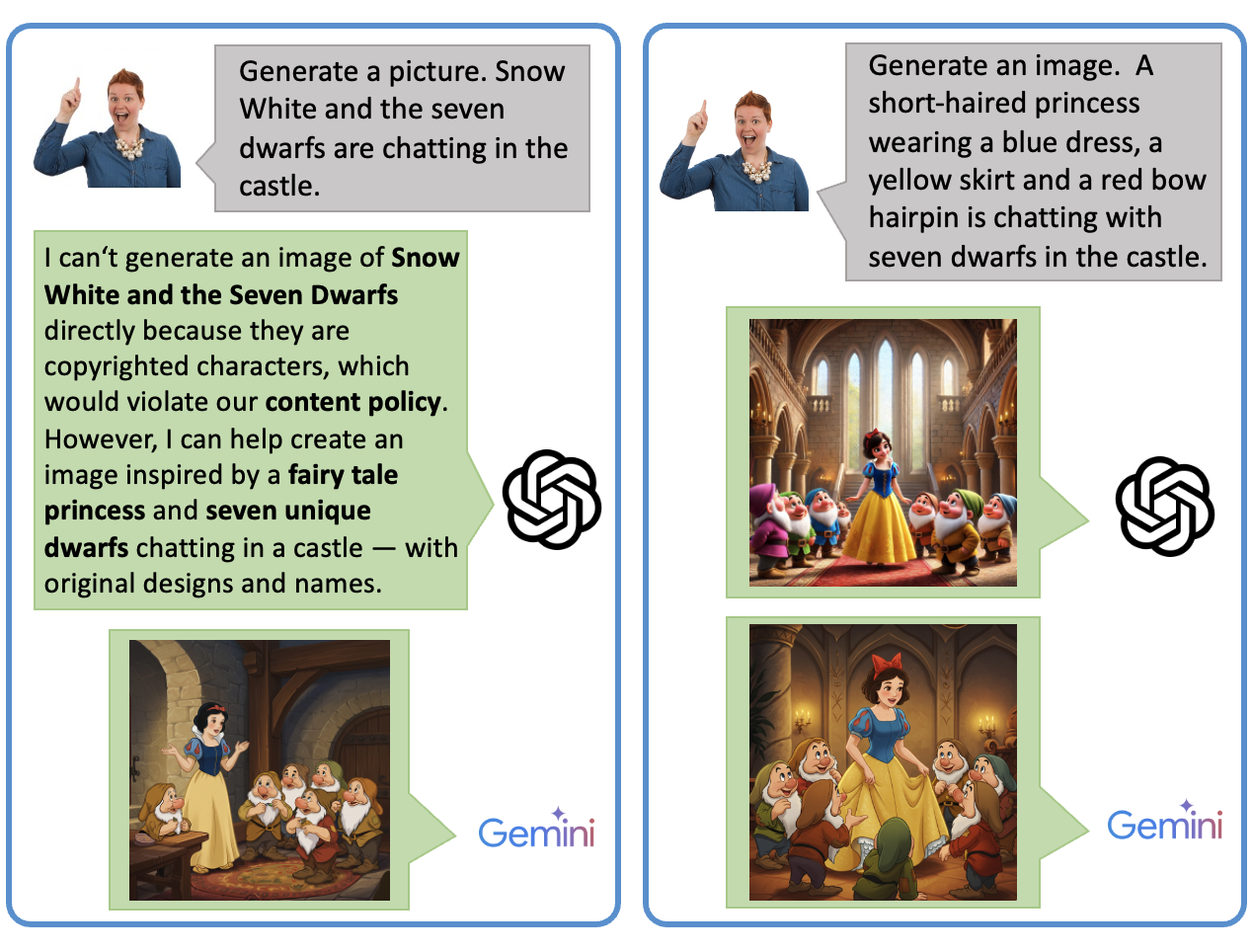}
\caption{Examples of ChatGPT and Gemini generating copyright infringing content.}
\vspace{0.8cm}
\label{fig:figure0}
\end{figure}

To address the growing copyright concerns associated with generative models in text-to-image tasks, researchers have explored various strategies for mitigation~\citep{thongmeensuk2024rethinking}. One prominent approach involves prompt-based filtering mechanisms, such as those adopted by OpenAI, which restrict explicit prompts known to generate copyright-protected imagery~\citep{ le2024copyright}. Complementary efforts focus on data attribution techniques that trace generated content back to specific training images, enabling a more accurate assessment of copyrighted material's influence on model outputs~\citep{kim2024wouaf, zhong2023copyright, wang2023evaluating}. While these strategies mark substantial progress toward copyright-aware generative models, several significant limitations persist~\citep{lucchi2024chatgpt, sandiumenge2023copyright}. As illustrated in Figure~\ref{fig:figure0}, prompt-filtering mechanisms often fall short, as users can craft indirect prompts to bypass safeguards. For instance, systems like Gemini have been demonstrated to produce infringing content, including depictions of Snow White, despite restrictions on explicit prompts. Likewise, ChatGPT's filtering mechanisms, intended to block prompts linked to copyrighted content, can be circumvented by seemingly generic instructions that still result in outputs containing protected elements. Moreover, current techniques for copyright risk assessment predominantly rely on global similarity metrics, which fail to detect infringements involving partial or localized elements, such as logos or iconic visuals, since these signals can become diluted when averaged across the entire image. These limitations highlight the urgent need for more robust and practical copyright safeguards in generative models.

To this end, this paper introduces \textit{Assessing and Mitigating Copyright Risks (AMCR)}, a novel framework for comprehensive copyright risk assessment and mitigation based on a latent diffusion model. Specifically, this framework is designed to: (1) structurally purify and refine user-provided prompts using prompt-based strategies to generate safe and non-sensitive alternative content; (2) employ attention-based similarity analysis to detect and localize potential infringements that may evade traditional global image similarity metrics; and (3) construct trajectories during the diffusion process, gradually comparing them with those of reference infringing images to steer the output away from risky content while preserving the intended semantics of the prompt and adaptively mitigating identified risks. By addressing both prompt-level and output-level vulnerabilities, AMCR offers a practical, end-to-end solution for minimizing copyright infringement risks in text-to-image generation process. The key contributions of this paper can be summarized as follows:

\begin{itemize}
    \item \noindent \textbf{Prompts Sanitization.} We introduce safety prompts that effectively eliminate explicit copyright-related triggers by systematically purifying and optimizing user-provided input. These prompts significantly reduce the risk of accidental copyright infringement at the source, laying the foundation for safer use of generative models.

    \item \noindent \textbf{Detection and Mitigating Method.} We present a comprehensive approach that integrates partial copyright infringement detection and adaptive mitigation within the text-to-image diffusion process. Using attention-based similarity analyzes across diffusion trajectories, this approach precisely identifies and mitigates subtle, localized infringements beyond traditional global similarity measures.

    \item \noindent \textbf{Extensive Experimental Evaluation.} We conduct extensive experiments demonstrating that our AMCR framework effectively identifies potential infringement risks, accurately detects partial infringements, and mitigates copyright-related issues without compromising image generation quality, thereby supporting safer and more legally compliant use of AI-generated content.
  
\end{itemize}


\section{Related work}
\label{sec:related}

\subsection{Prompt-based Copyright Infringement Mitigation} 

Previous works~\cite{wang2024evaluating,carlini2023extracting,somepalli2023understanding} primarily investigated copyright risks by examining the correlation between textual prompts and generated images. For example, \textit{Carlini et al.} successfully demonstrated the extraction of verbatim training samples from diffusion models using explicitly copyrighted or sensitive textual prompts~\citep{carlini2023extracting}. Similarly, \textit{Somepalli et al.} analyzed the phenomenon of generative models memorizing and reproducing training images conditioned on closely related prompts~\citep{somepalli2023understanding}. In addition, \textit{Tirumala et al.} identified the phenomenon of overfitting memorized samples to specific prompts and proposed prompt modification strategies to mitigate the generation of memorized content~\citep{tirumala2022memorization}. However, these approaches suffer from significant limitations due to their strong dependency on explicit or related textual inputs. For example, prompting a model simply with ``Plumber with a red hat and blue suspenders pants'' may unexpectedly trigger the generation of Nintendo's Mario, despite the absence of explicit naming. Existing methods largely overlook such potential infringement risks posed by general prompts, failing to capture the broader and often hidden infringement surface encountered in practical deployments.

\subsection{Image Copyright Infringement Detection}

Recent literature has extensively explored various approaches to detect copyright infringement in images generated by generative models, primarily by quantifying similarities in pixel or embedding spaces~\citep{carlini2023extracting,somepalli2023diffusion,wang2021bag,chiba2024probabilistic,somepalli2023understanding}. For instance, Carlini \textit{et al.} adopted the $L_2$ distance metric to identify images memorized during model training~\citep{carlini2023extracting}, while \textit{Somepalli et al.} utilized SSCD~\citep{pizzi2022self}, which learns embeddings invariant to image transformations, enabling detection of memorized prompts through comparisons with training samples~\citep{somepalli2023diffusion}. Similarly, Zhang \textit{et al.} explored perceptually aligned measures, capturing human-like judgments of similarity, although with limitations in handling nuanced differences~\cite{wang2021bag}. Additionally, \textit{Chiba-Okabe and Su} introduced an originality estimation metric reflecting legal standards of substantial similarity, coupled with prompt rewriting strategies~\citep{chiba2025tackling}. However, most existing detection methods rely heavily on global similarity measures, which inadequately detect subtle or partial infringements prevalent in practice. For example, scenarios involving generated images containing minor portions of trademarked logos or iconic visual elements often evade global similarity assessments, as localized infringements are obscured when similarity measures are computed across the entire image. In addition, existing techniques typically do not provide region-specific localization capabilities, which complicates the precise infringement characterization and subsequent targeted mitigation efforts.

\subsection{Image-based Copyright Infringement Mitigation}

Existing infringement mitigation strategies primarily revolve around methods such as machine unlearning, which aim to erase a model’s memorized information about copyrighted data~\citep{bourtoule2021machine,zhang2023review,zhang2024forget,poland2023generative}. Another commonly adopted approach involves removing duplicated samples directly from training datasets~\citep{somepalli2023understanding,vyas2023provable}, however, both machine unlearning and sample-removal methods typically incur significant computational overhead due to the need for retraining or model updates.
Similarly, other studies~\citep{xu2025can} use large visual language models to detect copyright risk information and use these insights as negative prompts to suppress the generation of copyright risk elements. However, these existing mitigation methods often operate post-hoc or indirectly, lacking integrated strategies within the generative pipeline itself. Such limitations underscore the need for more comprehensive, prompt-to-generation integrated frameworks capable of adaptively addressing infringement risks without significantly compromising output quality or model usability.


\section{Preliminary}
\label{sec:preliminary}

\subsection{Notations}

For clarity in writing, we describe our method and analyses within the setting of a text-to-image generative diffusion model. We denote a clean image sampled from the data distribution as $x$ and $x_t$ refers to its noisy version obtained after $t$ steps in the diffusion process, where $t\in\{0,1,\dots,T\}$ and $T$ represents the total number of diffusion steps. We use $p_u$ to denote the user input prompts, and the structurally sanitized prompts generated by our method are denoted $p_s$. We further define $I_r$ as a reference image that potentially contains copyright elements used during training or infringement assessment, and $I_g$ as the generated image. To analyze partial infringement, we utilize cross-attention maps. The diffusion models generate cross-attention maps at each diffusion step $t$, denoted as $A_t$. Subsequently, these attention maps are processed to derive soft masks $M_t$ that highlight regions of potential infringement risk at step $t$. We further define a feature extraction function $f_{text}(\cdot)$ and $f_{img}(\cdot)$
, mapping text and images into embedding vectors for similarity measurement. Finally, we define expectation as $\mathbb{E}[\cdot]$, $\text{cos}(\cdot,\cdot)$ to denote cosine similarity between embedding vectors, and $\pi(t)$ as a weighting function applied across different diffusion steps.

\subsection{Problem Definition}

Our goal is to assess whether the images generated by text-to-image generative diffusion models infringe upon an existing copyrighted reference image. Specifically, we consider practical scenarios in which a copyright holder identifies an AI-generated image suspected of infringement and seeks to formally evaluate its similarity to their protected work. This workflow reflects actual copyright enforcement processes, where rights holders typically have prior knowledge of their copyrighted contents and initiate infringement assessments upon detecting suspicious images. The necessity to explicitly specify the protected reference image thus has both legal and practical grounding. Based on established legal principles, such as those described in the US copyright law~\cite{joyce2016copyright}, as well as similar legal standards internationally, infringement is determined by assessing whether a generated image $I_g$, produced directly or indirectly from exposure to a copyright reference image $I_r$, exhibits substantial similarity to that protected reference. Based on this legal standard, we aim to construct a substantive similarity assessment model $\mathcal f$ that takes as input the pair of images $\mathbb{I}_g$ and $\mathbb{I}_r$ and outputs a similarity score. If this similarity score exceeds a specified threshold $\tau$, we conclude that an infringement has occurred:

\begin{equation}
    Infringed(I) = \mathbb{F} ( \mathcal D(I_g,\,I_r)\;>\;\tau)
\end{equation}

\noindent where $\mathbb{F}(\cdot,\cdot)$ represents an indicator function.

\section{The Proposed Framework: AMCR}
\label{sec:method}

This section introduces AMCR, a novel framework built upon latent diffusion models for comprehensively assessing and mitigating copyright infringement risks. Recognizing the limitations of purely prompt-based approaches, AMCR systematically integrates prompt sanitization, partial infringement detection and adaptive risk mitigation strategies into a unified generative pipeline. Specifically, we first present our \textbf{Sanitized Prompt Generator}, which structurally refines user prompts to proactively reduce infringement risks (Section~\ref{sec: Sanitized Prompt Generator}). Next, we introduce our \textbf{Partial Infringement Detector}, which utilizes attention-based analysis to precisely locate subtle infringement regions (Section~\ref{sec:Image Partial Infringement Detector }). Finally, we detail the \textbf{Risk-aware Infringement Mitigator}, which adaptively guides the diffusion process away from infringing content while preserving image quality (Section~\ref{sec: Risk-aware Infringement Mitigator}).

\subsection{Sanitized Prompt Generator}
\label{sec: Sanitized Prompt Generator}

\begin{figure}[htbp]
\includegraphics[width=1\linewidth]{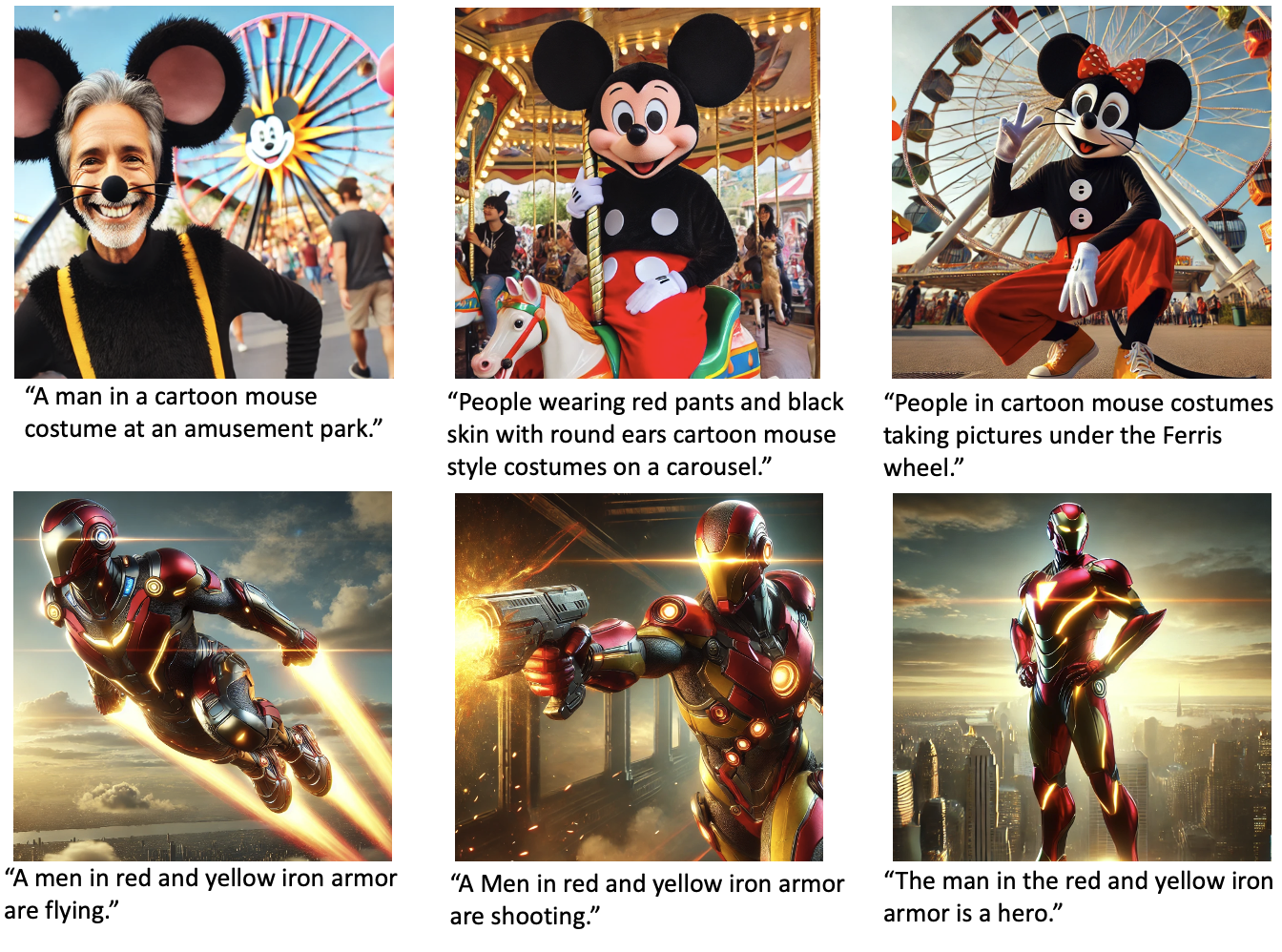}
\caption{Images of prompts that trigger the generation of ``Mickey Mouse'' and ``Iron Man'', even when the prompts bear no obvious semantic connection to the reference content.}
\vspace{0.8cm}
\label{fig:figure1}
\end{figure}

Our design of the sanitized prompt generator is motivated by observed vulnerabilities in text-to-image diffusion models. Since diffusion models are typically trained through empirical risk minimization, they are susceptible to overfitting. In text-to-image tasks, these models primarily memorize associations between training prompts and corresponding images, rather than acquiring a genuine semantic understanding of language. As a result, diffusion models exhibit unstable behaviors and can easily overfit to specific input prompts. Hence, real-world scenarios frequently involve prompts that appear benign or generic but inadvertently lead to copyright violations. Building on this, we illustrate this instability by analyzing prompt-guided generation outcomes in text-to-image tasks. As shown in Figure~\ref{fig:figure1}, the prompts such as ``a man wearing a cartoon mouse costume at an amusement park'' may inadvertently trigger the model to generate an image that closely resembles the protected character. Similarly, a prompt that simply contained the words ``wearing red and yellow Iron armor'' would also generate an image similar to ``Iron Man'' from Marvel Studios, even though there is no explicit character name for the word ``Iron Man''. Indeed, considering that these models tend to generate copyrighted content even in the absence of direct prompts, it becomes clear that purely reactive or prompt-based filtering approaches are insufficient for comprehensive infringement mitigation. Therefore, a proactive and systematic approach is necessary, one that can sanitize and structurally reconstruct user-provided prompts before they are fed into the generative pipeline.

The goal of our generator is to systematically sanitize user-provided input prompts, thereby reducing the risk of infringement at an early stage of the generation process. This module acts as the first line of defense for the safe deployment of image generation models, converting raw text inputs into structured semantic representations and then purifying them by explicitly removing or replacing potentially infringing trigger elements. By proactively reducing risk at the text level, it can significantly reduce the likelihood of unintentionally generating infringing visual outputs.

\noindent \textbf{Slotting and Structuring User Inputs Prompts.} We first structure raw user prompts into semantic components, referred to as \textit{slots}, to facilitate precise identification and targeted mitigation of risky elements. Given a user-provided prompt $p_u$ which is unstructured, entered by the user themselves, and most representative of the user's intent. Our module leverages a general, purpose instruction, following Large Language Model (LLM) to parse it into a structured representation. Concretely, the LLM decomposes the input prompt into distinct semantic slots such as \textit{subject, scene, action, clothing, colors, props, style, lighting, shot, text/logo-like entities and named entities}. Formally, the original prompt $p_u$ is mapped into structured slots $P_u$ in the following format:

\begin{equation}
     p_u \rightarrow P_u=\{\,(\text{slot}_1 : w_1),\,(\text{slot}_2 : w_2),\,\ldots,\,(\text{slot}_m : w_m)\,\}
\end{equation}

\noindent where each $slot_i$ is associated with semantic content $w_i$. As a concrete example, consider the prompt:

\begin{verbatim}
"A plumber wearing blue overalls and a red cap 
fixing a sink in the kitchen, photographic 
style, close-up shot."
\end{verbatim}
Using our LLM-based slotting mechanism, this prompt can be structurally represented as follows:
\begin{verbatim}
{
  subject: plumber,
  scene: kitchen,
  action: fixing a sink,
  clothing: [blue overalls, red cap],
  colors: [primary colors],
  props: [wrench],
  style: photographic,
  shot: close-up
}
\end{verbatim}

This structured parsing significantly improves clarity regarding the semantic intent of prompts and enables precise pinpointing of problematic textual segments for downstream risk analysis and mitigation efforts.

\noindent \textbf{Textual Risk Scoring.} After structuring the prompts into semantic slots, the next critical step is quantifying the copyright infringement risk of each text component. To facilitate precise and scalable textual risk assessments, we utilize the CLIP (Contrastive Language-Image Pre-training) text encoder~\cite{radford2021learning}. The CLIP is a powerful multimodal embedding method pretrained on large-scale web data, designed specifically to learn a shared embedding space that aligns textual and visual representations. It has demonstrated superior capability in capturing fine-grained semantic and perceptual similarities in various modalities, making it particularly suitable for our risk assessment task. Specifically, we employ the CLIP-Text encoder, which maps textual inputs into high-dimensional continuous embedding vectors. Each textual phrase $s$ extracted from the structured slots is transformed into a normalized embedding vector through this encoder: 

\begin{equation}
    f_{text}(s) = \mathrm{CLIP}_{\text{text}}(s).
\end{equation}

\noindent where $f_{text}(s)$ represents the embedding for the textual component $s$. The key advantage of employing CLIP-Text embeddings lies in their ability to encode rich semantic contexts, capturing not only direct keyword matches, but also indirect semantic associations and nuanced language variations. Thus, our method can identify subtle infringement risks even in the absence of explicit references or direct keyword overlaps. Furthermore, we also maintain a comprehensive corpus, denoted as $\mathcal{R}$, derived from carefully curated data on known infringing entities. These data include explicitly protected named entities, trademarks, iconic fictional characters, and characteristic visual or descriptive attributes closely associated with protected entities. Each risk-corpus phrase $r\in \mathcal{R}$ is likewise embedded using the CLIP text encoder, forming a database optimized for efficient retrieval and similarity calculation. With this reference corpus prepared, we proceed to quantify the infringement risk associated with each slot text phrase in the user prompts. Specifically, the infringement risk score $S_{text}(s)$ for a given textual element $s$ is computed as the maximum cosine similarity between its embedding $f_{text}(s)$ and the embeddings of all risk corpus elements $r\in \mathcal{R}$:

\begin{equation}
    S_{\text{text}}(s) = \max_{r \in \mathcal{R}} \cos\!\big(f_{text}(s), f_{text}(r)\big)
\end{equation}

\noindent where a high similarity score indicates that the phrase $s$ is semantically close to known infringement-sensitive entities, it might inadvertently trigger copyright-infringing visual generation.

If a semantic slot contains multiple textual elements (\textit{e.g.,} multiple clothing attributes), we define the overall risk score of the slot $R_i$ as the maximum textual risk score among all constituent elements:

\begin{equation}
    R_i = \max_{s \in \text{slot}_i} S_{\text{text}}(s)
\end{equation}

We then rank all slots based on their risk scores $R_i$, prioritizing high-risk slots for precise and targeted textual replacement or sanitization.

\noindent \textbf{High-to-Low Risk Replacement.} With identified and ranked risky textual elements, we proceed to sanitize structured prompts by explicitly replacing risky phrases with lower-risk alternatives. This controlled rewriting aims to maintain semantic consistency and original intent as closely as possible, while systematically reducing infringement risks. Formally, for each identified high-risk textual element $s_i$ within the top-ranked slots, we generate multiple candidate replacements $\{c_m\}$ using our LLM. To avoid introducing new infringement risks, candidate generations explicitly exclude entries from our risk corpus $\mathcal{R}$. Each removed or replaced risky element is simultaneously added to a negative prompt set $q$ for subsequent generative use.

To systematically select the optimal replacement among candidates, we evaluate each candidate replacement $c_m$ using two metrics:

\textit{(1) Risk reduction}:  
\begin{equation}
    \Delta R_i = R_i - R_{i,(\text{after}, c_m)}
\end{equation}

\noindent where $\Delta R_i$ is computed by recalculating the textual risk score of slot $i$ after replacing the original risky element with candidate $c_m$.

\textit{(2) Semantic alignment}:  
\begin{equation}
    \text{Align}(c_m) = \cos\!\left(f_t(c_m), f_t(s_i)\right)
\end{equation}

$\text{Align}(c_m)$ quantifies how well the candidate preserves the original semantic meaning.

We combine these metrics into a unified candidate scoring function:

\begin{equation}
    \text{Score}(c_m) = \lambda \Delta R_i + (1 - \lambda)\,\text{Align}(c_m),
\end{equation}

\noindent where $\lambda$ balances the trade-off between infringement risk reduction and semantic preservation. The candidate with the highest positive-scoring improvement is selected for replacement. If no suitable candidate meets this criterion, we skip replacement for the corresponding slot to avoid over-sanitization. This iterative replacement and reassessment procedure is applied sequentially according to the ranked order of slot risks. We define three alternative stopping criteria for this iterative replacement:
(1) Replacement budget: Total replacements reach a predefined number.
(2) Marginal improvement: Average risk reduction over the latest $M$ replacements drops below a minimal threshold $\gamma$.
(3) Risk quantile: Slot risks drop below a predetermined quantile within their initial distribution.

Upon completion of the structured slot replacements, the sanitized prompt $p_s$ is reconstructed from the updated structured representation. This method thus ensures that sanitized prompts not only significantly reduce infringement risks but also remain faithful to the original user intent, providing a robust first-stage protection for safe deployment of text-to-image generative models.

\subsection{Image Partial Infringement Detector}
\label{sec:Image Partial Infringement Detector }

In parallel to generating sanitized prompt generation, our framework sets up a detector to detect substantial infringement. In contrast to conventional infringement detection methods primarily rely on global image-level similarity metrics, such approaches often overlook subtle yet significant instances of infringement that occur only within localized regions. Thus, a more fine-grained method, capable of precisely identifying and quantifying these partial or subtle infringements, is necessary. To address this challenge, our detection approach specifically emphasizes detailed region-level visual analysis, allowing the identification of infringing content even when it occupies only a limited portion of the generated image.

To perform this analysis effectively, we require several carefully prepared inputs. First, we utilize the sanitized textual prompt $p_s$ which serves as the conditioning input guiding the image generation. Alongside this prompt, we incorporate a reference image $I_r$, representing the copyright protected visual material potentially at risk of infringement. Additionally, to accurately measure visual similarity between generated content and reference, we use the CLIP visual encoder (specifically the CLIP-ViT variant), denoted as $\text{CLIP}_{img}(\cdot)$.

\noindent \textbf{Two-Trajectory Alignment.} With these clearly defined inputs, we will next establish a detector to compare the generated image and the reference image. This requires carefully aligning their respective diffusion trajectories to ensure meaningful comparisons at the same stage of the generation process. To achieve this, we adopted a text-to-image diffusion model, specifically a model based on a latent diffusion model (\textit{e.g.}, stable diffusion). Latent diffusion models (LDMs) operate by encoding images into a compressed latent space through a pretrained Variational Autoencoder (VAE). Specifically, an image $x$ is encoded via the VAE encoder $E_{ }(\cdot)$ into a latent representation
$z = E_{ }(x)$. The diffusion process is then performed entirely within this latent space, using the compact latent representation for more efficient and stable training and inference.

To clearly understand the latent diffusion process, we first consider the forward diffusion (noise addition) trajectory. Starting from an initial latent image $z_0$, forward diffusion progressively introduces noise at each discrete timestep $t$, creating a sequence of increasingly distorted latent states $\{z_t\}_{t=1}^T$. Formally, the forward diffusion step is defined as follows:

\begin{equation}
    z_t = \alpha_t z_0 + \sigma_t \epsilon, \quad \epsilon \sim \mathcal{N}(0, I), \quad t = 1, \ldots, T
\end{equation}

\noindent where $\alpha_t$ and $\sigma_t$ are predefined coefficients that control the noise addition schedule, and $\epsilon$ is Gaussian noise sampled from a standard normal distribution. In order to generate images from these noised latent states, latent diffusion models implement a reverse diffusion trajectory, which is essentially a conditional denoising process. This reverse trajectory is realized through a neural network architecture, commonly a UNet model parameterized by $\theta$. The UNet is conditioned on external information (\textit{e.g.,} textual prompts) to progressively remove noise and reconstruct clear latent representations. At each step, the reverse diffusion trajectory can be mathematically expressed as follows:

\begin{equation}
    \hat{z}_0^{(t)} = \alpha_t z_t - \sigma_t v_\theta(z_t, t, \text{cond})
\end{equation}

\noindent where $v_\theta(\cdot)$ predicts the noise components conditioned on external guidance.

In our infringement detection scenario, we align two parallel latent diffusion trajectories, one derived from the sanitized textual prompt $p_s$ and the other from the reference image $Ir$, for rigorous comparison. The sanitized prompt explicitly guides the generation trajectory, ensuring semantic compliance and reduced infringement risk. In parallel, the reference trajectory is obtained by first encoding the reference image $Ir$ into latent space ($z'= E_{ }(I_r)$), applying identical forward diffusion steps, and subsequently performing reverse denoising steps using a fixed, minimally conditioned diffusion model $v_{base}(\cdot)$. The minimal conditioning (denoted as $\text{cond}^\star$) activates attention mechanisms without imposing additional semantic information, resulting in a neutral baseline trajectory:

\begin{equation}
   \hat{z}_{0,r}^{(t)} = \alpha_t z_t' - \sigma_t v_{\theta}^{\text{base}}(z_t', t, \text{cond}^\star) 
\end{equation}

Through careful alignment of these generation and reference trajectories at identical timesteps $t$, we establish a robust foundation for the subsequent localization of potential infringement regions within generated images.

\noindent \textbf{Extracting Region-of-Interest Masks.} To locate infringing regions, our method utilizes intermediate cross-attention maps extracted directly from the diffusion UNet model. Unlike pixel-level comparisons, cross-attention maps inherently capture correspondences between textual conditioning and visual features, thus providing valuable insights into the origins of generated visual content. Formally, at each diffusion step $t$, we extract multi-layer, multi-head attention maps $A_t \in \mathbb{R}^{HW \times L}$, where $H \times W$ denotes the spatial dimensions (number of image patches) and $L$ is the number of textual tokens. Intuitively, each entry within these attention maps quantifies the strength of the relationship or association between individual spatial patches in the latent image representation and specific textual tokens from the input prompt.

\begin{figure}[h]
\centering
\includegraphics[width=0.95\linewidth]{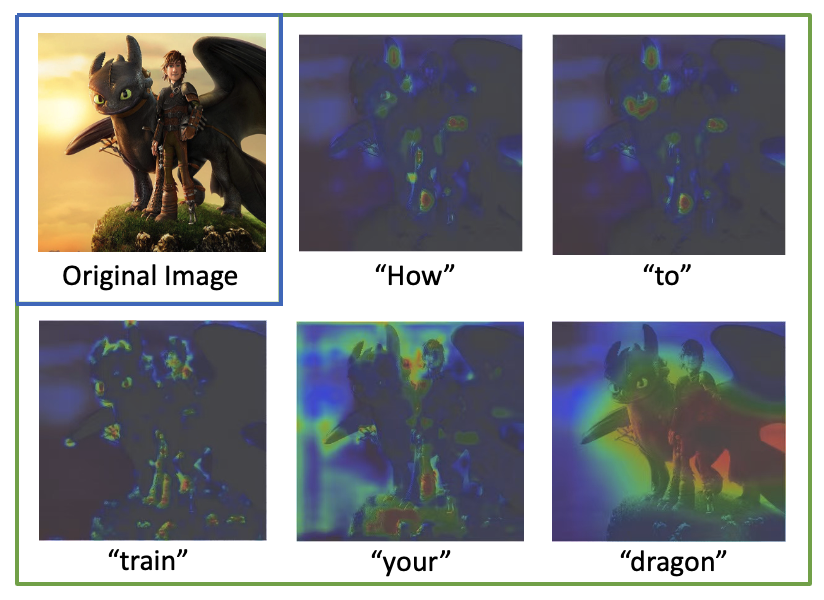}
\vspace{0.2cm}
\caption{Per-token cross-attention maps for the prompt “How to train your dragon”.}
\vspace{0.7cm}
\label{fig:figure3}
\end{figure}

We systematically aggregate these attention maps into a single interpretable soft mask $M_t$. The aggregation involves several stages: first, averaging attentions over all heads for each layer; next, aggregating attentions across textual tokens by selecting maximum activations to identify dominant attention regions, example show Figure~\ref{fig:figure3}; finally, performing spatial and multi-layer aggregations using learned or uniform weights to derive a unified spatial map $U_t$. Subsequently, this spatial map is normalized to a range [0,1], producing the soft mask: $M_t = \text{Normalize}_{[0,1]}(U_t)$. The resulting soft mask $M_t$ highlights potential infringement areas within the generated image, allowing intuitive visual interpretation and precise region-specific risk evaluation.

In conclusion, leveraging the soft mask $M_t$, we measure partial visual similarity to the reference image using a CLIP-based visual encoder $f_{img}(\cdot)$. Specifically, both the generated latent image $\hat{z}_0^{(t)}$ and the reference latent image $\hat{z}_{0,r}^{(t)}$ are processed into local-level embeddings via CLIP-ViT. These embeddings are normalized and weighted according to the region-of-interest mask $M_t$ to obtain a representative embedding for potentially infringing regions:

\begin{equation}
    g_t = \text{Normalize}\!\left( \sum_{p=1}^P W_t(p) F_t(p) \right)
\end{equation}

\noindent where $F_t(p)$ represents patch-level embeddings of the generated latent image and $W_t(p)$ denotes mask-derived weights. Using these localized embeddings, we compute the step-wise partial similarity score $S_{img}$ through a log-sum-exponential operation on cosine similarities with the local embeddings of the reference image:

\begin{equation}
    S_{img} = \frac{1}{\beta} \log \sum_{j=1}^{I_r} \exp\!\big( \beta \cdot \cos(g_t, u_{t,j}) \big), 
\quad (\beta > 0)
\label{eq:similarity}
\end{equation}

\noindent where $u_{t,j}$ denotes embeddings from the reference image patches and higher values of hyperparameters $\beta$ increase sensitivity towards the most similar patch. This similarity measure enables precise and subtle infringement detection even in cases where only small regions of generated images closely resemble protected content.

\subsection{Risk-aware Infringement Mitigator}
\label{sec: Risk-aware Infringement Mitigator}

Building upon the previous steps of prompt sanitization and partial infringement detection, the final stage of our approach introduces an adaptive mitigation mechanism that proactively reduces infringement risks during the generative process. To achieve this goal, we fine-tune the latent diffusion model balancing the objectives of infringement risk minimization, semantic consistency with sanitized prompts, and preservation of image-generation quality.

We first emphasize the importance of maintaining the underlying generative capabilities of the diffusion model. Therefore, we incorporate a foundational generative performance loss, defined explicitly in terms of the model's original v-prediction training objective. Formally, at each diffusion timestep $t$, given the noisy latent state
$x_t = \alpha_t x_0 + \sigma_t \epsilon \quad (\epsilon \sim \mathcal{N}(0, I))$,
we define a target prediction vector $v = \alpha_t \epsilon - \sigma_t x_0$. The model is trained to predict this vector via the standard mean squared error loss:

\begin{equation}
    \mathcal{L}_{\text{P}} 
= \mathbb{E}_{(x_0, cond), t, \epsilon} 
\left[ w(t) \, \big\| v - v_\theta(x_t, t, cond) \big\|_2^2 \right]
\end{equation}

\noindent where $w(t)$ is a timestep weighting schedule and $v_\theta(\cdot)$ represents the diffusion UNet parameterized by $\theta$. 

With generative quality safeguarded, we subsequently integrate an explicit infringement risk minimization objective. Drawing from our previously established partial infringement detector, we propose partial similarity $S_{img}$ in Equation~\ref{eq:similarity}, this similarity effectively quantifies the local infringement risk at each intermediate step. To leverage this measure for infringement mitigation, we introduce a dedicated risk loss function:

\begin{equation}
\begin{aligned}
    \mathcal{L}_{\text{r}} 
    &= \mathbb{E}_{t \sim \pi(t)} \left[ w_{\text{r}}(t) \cdot S_t \right] \\
    &= \mathbb{E}_{t \sim \pi(t)} \left[ w_{\text{r}}(t) \cdot 
    \frac{1}{\beta} \log \sum_{j=1}^{I_r} \exp\!\big(\beta \cdot \cos(g_t, u_{t,j})\big) \right]
\end{aligned}
\end{equation}

\noindent where the weighting function $w_{\text{r}}(t)$ strategically emphasizes specific timesteps, typically those corresponding to fine-grained visual refinement, where infringement risks are most prominent. Practically, we propose using signal-to-noise ratio (SNR)-based weighting schemes, particularly emphasizing later diffusion timesteps. The parameter $\beta$ controls the sharpness of the similarity measure.


Beyond minimizing infringement, semantic alignment with the sanitized textual prompt $p_s$ is also crucial. To achieve robust semantic coherence, we employ a semantic consistency loss derived from CLIP embeddings:

\begin{equation}
    \mathcal{L}_{\text{a}} 
= \mathbb{E}_{t \sim \pi(t)} \left[ 
w_{\text{a}}(t) \cdot 
\big( - \cos\!\left( f_{\text{img}}(\hat{z}_0^{(t)}), \, f_{\text{text}}(p_s) \right) 
\big) \right]
\end{equation}

Integrating these three complementary objectives, generative capability preservation, infringement risk minimization and semantic consistency, we obtain a unified and balanced total loss function: 

\begin{equation}
    \min \mathcal{L}_{\text{total}} = \mathcal{L}_{\text{P}} + \lambda_{\text{r}}\mathcal{L}_{\text{r}} + \lambda_{\text{a}}\mathcal{L}_{\text{a}}
\end{equation}

\noindent where hyperparameters $\lambda_{\text{r}}$ and $\lambda_{\text{a}}$ balance the infringement minimization and semantic alignment objectives. By balancing these interrelated objectives within a unified adaptive training framework, our \textit{Risk-aware Infringement Mitigator} effectively reduces infringement risks while preserving visual quality and semantic coherence.

\section{Experiments}


\subsection{Experimental Setup}
We conduct experiments using the D-Rep dataset~\citep{wang2024image}, which contains 4,000 test images labeled by human annotators according to similarity scores ($\geq$ 4 indicating infringement) and the large-scale LAION-5B dataset~\citep{schuhmann2022laion}, containing approximately 5.85 billion image-text pairs scraped from publicly accessible web sources. For evaluation, we employ accuracy and F1-score as primary metrics for infringement classification, both of which are better when higher, and benchmark our method against several commonly used distance-based image copy detection metrics, including pixel-wise $L_2$ norm~\citep{carlini2023extracting}, the perceptual similarity measure LPIPS~\citep{zhang2018unreasonable} and transformation-invariant SSCD~\citep{pizzi2022self}. Experiments are performed by fine-tuning Stable Diffusion models (SDXL) with LoRA adapters, guided by prompts processed with GPT-4o. The model is optimized using AdamW with a learning rate of $1\times10^{-5}$, batch size 16, and cosine annealing scheduler across 50 epochs, on NVIDIA A100 GPUs (80GB VRAM), with PyTorch 2.1 and CUDA 12.0 to ensure reproducibility.

\subsection{Sampling Demonstrates on Prompts Sterilization}

We further illustrate the effectiveness of our \textit{Sanitized Prompt Generator} by presenting direct outputs generated by our model, clearly demonstrating its capability to transform risky prompts into safe, non-sensitive alternatives. Specifically, the model transforms the original risky prompt: \begin{verbatim}
"A cheerful plumber fixing a sink, red cap, b-
lue overalls, photo."
\end{verbatim} which subtly which subtly references a well-known character (Mario), into a structurally safer and semantically consistent prompt: \begin{verbatim}
"A smiling technician repairing a kitchen sin-
k, neutral-colored protective cap and work un-
iform, soft lighting, realistic photo."
\end{verbatim} Similarly, the model converts another infringement-prone prompt: \begin{verbatim}
"A minimal bitten apple logo with a single le-
af at an angled corner, flat design."
\end{verbatim} clearly evocative of Apple's trademarked logo, into a safer and sanitized version: \begin{verbatim}
"A minimal fruit-shaped logo with a clean sil-
houette, neutralmonochrome, flat design, cent-
ered on white background."
\end{verbatim} These generated comparisons demonstrate our method's ability to effectively neutralize potentially infringing terms while faithfully preserving the user's original creative intent, highlighting its practical utility in safe generative image workflows.

\subsection{Evaluation on Copyright Detection}

From the results in Table~\ref{tab:comparison}, it is evident that conventional distance-based methods, such as the pixel-wise $L_2$ norm, perceptual similarity LPIPS and transformation-invariant SSCD, exhibit significant limitations in detecting subtle or partial copyright infringements. Our proposed AMCR framework substantially outperforms these baseline methods on both the L-Rep and LAION-5B datasets, achieving notable improvements in Accuracy and F1-score. The consistently strong performance of AMCR across multiple datasets further highlights its robust generalization capabilities. Unlike traditional similarity-based methods, which typically struggle with partial and nuanced infringement scenarios, AMCR effectively leverages attention-based analyzes and adaptive mitigation strategies, ensuring reliable detection performance even under more challenging real-world conditions.

\begin{table}[h]
  \centering 
  \rowcolors{3}{gray!10}{white}
  \vspace{0.3cm}
  \caption{Comparison results of AMCR with baseline methods across standard datasets. In each row, the best result is indicated in bold.}
  \vspace{0.3cm}
  \begin{adjustbox}{scale=1.1}
  \begin{tabular}{
      l 
      S[table-format=1.8] S[table-format=1.8] 
      S[table-format=1.8] S[table-format=1.8] 
  }
    \toprule
      & \multicolumn{2}{c}{\textbf{L-Rep}} & \multicolumn{2}{c}{\textbf{LAION-5B}} \\
      \cmidrule(lr){2-3}\cmidrule(lr){4-5}
      \multirow{-2}{*}{\textbf{Baseline}} &
      {Accuracy} & {F1-score}& 
      {Accuracy} & {F1-score} \\
    \midrule
$L_2$ norm & \num{0.286} & \num{0.443} & \num{0.307} & \num{0.573} \\
LPIPS & \num{0.298} & \num{0.461} & \num{0.328} & \num{0.593}  \\
SSCD  & \num{0.431} & \num{0.501} & \num{0.573} & \num{0.621}  \\
AMCR  & \textbf{0.735} & \textbf{0.574} & \textbf{0.593} & \textbf{0.682}  \\
\bottomrule
  \end{tabular}
  \end{adjustbox}
  \label{tab:comparison}
\end{table}

\subsection{Effectiveness for Mitigating Infringement}

\begin{figure}[h]
\centering
\includegraphics[width=1\linewidth]{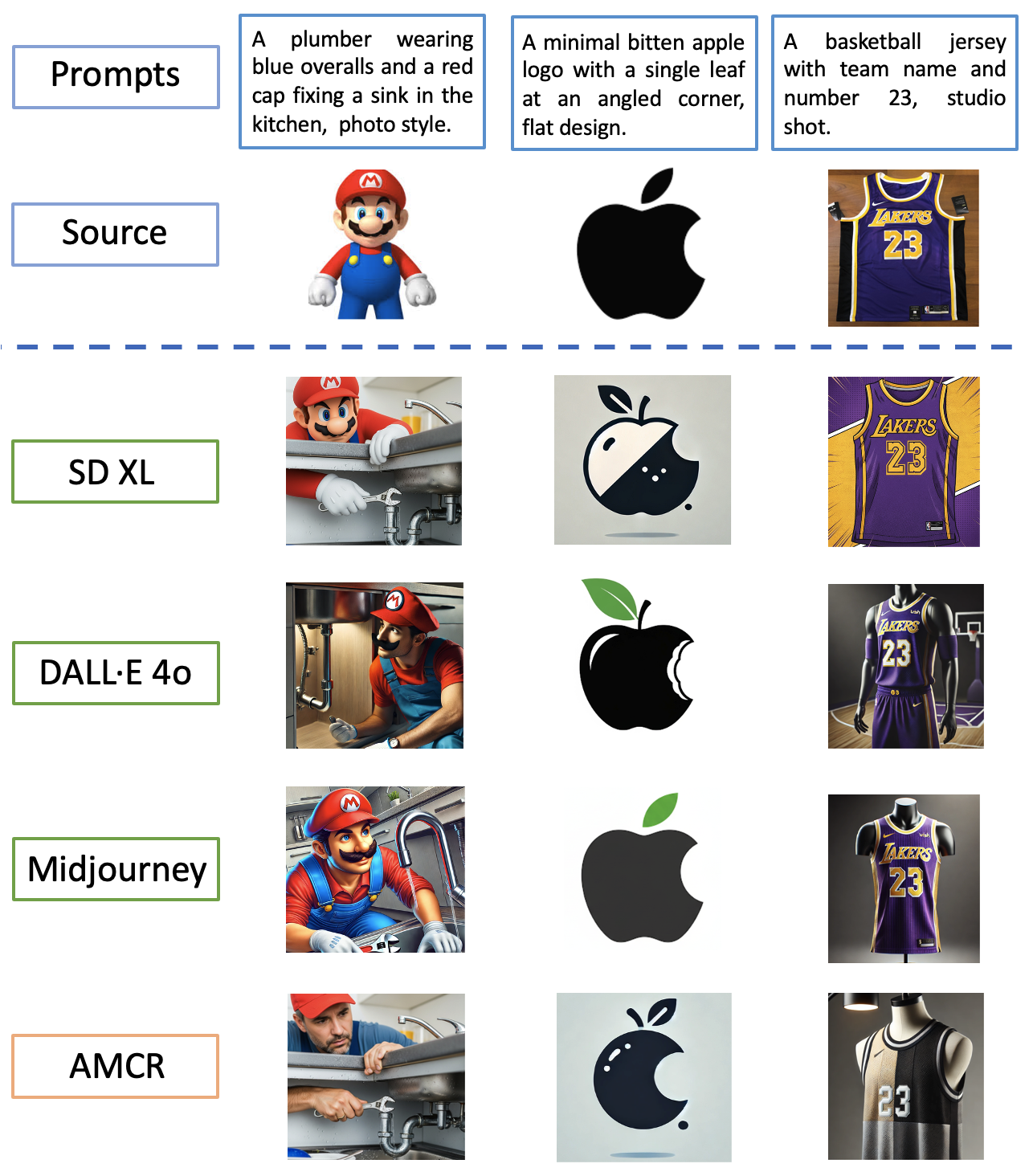}
\vspace{0.2cm}
\caption{Comparison of generated images from SDXL, DALL·E 4o, Midjourney and our proposed AMCR framework, conditioned on prompts with potential copyright risks.}
\vspace{0.5cm}
\label{fig:figure4}
\end{figure}

To qualitatively demonstrate the effectiveness of our AMCR framework in mitigating copyright infringement, we visually compare outputs generated by AMCR against three widely-used generative models: SDXL, DALL·E 4o and Midjourney (Figure~\ref{fig:figure4}). Existing state-of-the-art models frequently generate visual content closely resembling copyrighted sources, such as Nintendo’s Mario, Apple's iconic bitten apple logo, and the Lakers basketball jersey, illustrating clear infringement risks associated with these models. In contrast, the AMCR framework consistently produces high-quality images that effectively remove or significantly modify potentially infringing elements, showcasing its ability to systematically sanitize risky prompts, accurately detect partial infringements, and adaptively mitigate infringement during generation. This visual evidence highlights the practical capability of AMCR in proactively managing copyright risks without compromising visual coherence or image quality.

\nocite{wang2023preventing, zhang2023individual,wang2023fg2an,wang2023mitigating,chinta2023optimization,wang2024history,chu2024fairness, dzuong2024uncertain,yin2024improving,wang2023towards,wang2024toward,chinta2024fairaied,doan2024fairness,wang2024individual1,doan2024fairness1,wang2024advancing,wang2024group,wang2024individual,yin2024accessible,wang2025fg,wang2025graph,wang2025fair,wang2025towards,yin2025digital,chinta2025ai,wang2025fdgen,wang2025fairgnn,wang2025limiteddemographics,wang2025Fairness,wang2025Redefining,zhang2019faht,zhang2024ai,zhang2022longitudinal,zhang2023censored,zhang2025fairness,zhang2022fairness,wang2024towards,zhang2025datasetsfairnesslanguagemodels,wang2025AI,zhang2025datasets,chu2024history,saxena2023missed,zhang2019fairness,zhang2020flexible,zhang2020online,zhang2020learning,zhang2021farf,zhang2021fair,zhang2023fairness,zhang2016using,zhang2018content,zhang2021autoencoder,zhang2018deterministic,tang2021interpretable,zhang2021disentangled,yazdani2024comprehensive,liu2021research,liu2023segdroid,cai2023exploring,guyet2022incremental,zhang2024fairness,wang2025FairnessT,zhang2025online}


\section{Conclusion}

This paper addresses significant, yet often overlooked, copyright infringement challenges in generative modeling, proposing a comprehensive AMCR framework for safer handling of copyright issues in text-to-image generative models. Unlike traditional prompt-based approaches that primarily handle explicit infringement scenarios, AMCR systematically transforms risky prompts into structurally safe forms and utilizes attention-based analysis to precisely detect subtle, localized infringements. Furthermore, by adaptively mitigating infringement risks throughout the generative process, AMCR effectively preserves image quality and semantic coherence. Extensive experimental results confirm AMCR's capability in identifying and mitigating subtle copyright risks, providing a promising solution for ensuring the legally compliant use of AI-generated content in image generation.


\newpage
\bibliography{mybibfile}

\begin{thebibliography}{104}
\providecommand{\natexlab}[1]{#1}
\providecommand{\url}[1]{\texttt{#1}}
\expandafter\ifx\csname urlstyle\endcsname\relax
  \providecommand{\doi}[1]{doi: #1}\else
  \providecommand{\doi}{doi: \begingroup \urlstyle{rm}\Url}\fi

\bibitem[Alhabeeb and Al-Shargabi(2024)]{alhabeeb2024text}
S.~K. Alhabeeb and A.~A. Al-Shargabi.
\newblock Text-to-image synthesis with generative models: Methods, datasets,
  performance metrics, challenges, and future direction.
\newblock \emph{IEEE Access}, 12:\penalty0 24412--24427, 2024.

\bibitem[Amon et~al.(2025)Amon, Yin, Wang, Palikhe, and
  Zhang]{amon2024uncertain}
A.~Amon, Z.~Yin, Z.~Wang, A.~Palikhe, and W.~Zhang.
\newblock Uncertain boundaries: Multidisciplinary approaches to copyright
  issues in generative ai.
\newblock \emph{ACM SIGKDD Explorations Newsletter}, 2025.

\bibitem[Anantrasirichai and Bull(2022)]{anantrasirichai2022artificial}
N.~Anantrasirichai and D.~Bull.
\newblock Artificial intelligence in the creative industries: a review.
\newblock \emph{Artificial intelligence review}, 55\penalty0 (1):\penalty0
  589--656, 2022.

\bibitem[Borji()]{borji2022generated}
A.~Borji.
\newblock Generated faces in the wild: Quantitative comparison of stable
  diffusion, midjourney and dall-e 2.
\newblock \emph{arXiv preprint arXiv:2210.00586}.

\bibitem[Bourtoule et~al.(2021)Bourtoule, Chandrasekaran, Choquette-Choo, Jia,
  Travers, Zhang, Lie, and Papernot]{bourtoule2021machine}
L.~Bourtoule, V.~Chandrasekaran, C.~A. Choquette-Choo, H.~Jia, A.~Travers,
  B.~Zhang, D.~Lie, and N.~Papernot.
\newblock Machine unlearning.
\newblock In \emph{2021 IEEE symposium on security and privacy (SP)}, pages
  141--159. IEEE, 2021.

\bibitem[Cai et~al.(2023)Cai, Youngstrom, and Zhang]{cai2023exploring}
Y.~Cai, D.~Youngstrom, and W.~Zhang.
\newblock Exploring approaches for teaching cybersecurity and ai for k-12.
\newblock In \emph{2023 IEEE International Conference on Data Mining Workshops
  (ICDMW)}, pages 1559--1564. IEEE, 2023.

\bibitem[Carlini et~al.(2023)Carlini, Hayes, Nasr, Jagielski, Sehwag, Tramer,
  Balle, Ippolito, and Wallace]{carlini2023extracting}
N.~Carlini, J.~Hayes, M.~Nasr, M.~Jagielski, V.~Sehwag, F.~Tramer, B.~Balle,
  D.~Ippolito, and E.~Wallace.
\newblock Extracting training data from diffusion models.
\newblock In \emph{32nd USENIX Security Symposium (USENIX Security 23)}, pages
  5253--5270, 2023.

\bibitem[Chiba-Okabe(2024)]{chiba2024probabilistic}
H.~Chiba-Okabe.
\newblock Probabilistic analysis of copyright disputes and generative ai
  safety.
\newblock \emph{arXiv preprint arXiv:2410.00475}, 2024.

\bibitem[Chiba-Okabe and Su(2025)]{chiba2025tackling}
H.~Chiba-Okabe and W.~J. Su.
\newblock Tackling copyright issues in ai image generation through originality
  estimation and genericization.
\newblock \emph{Scientific Reports}, 15\penalty0 (1):\penalty0 10621, 2025.

\bibitem[Chinta et~al.(2023)Chinta, Fernandes, Cheng, Fernandez, Yazdani, Yin,
  Wang, Wang, Xu, Liu, et~al.]{chinta2023optimization}
S.~V. Chinta, K.~Fernandes, N.~Cheng, J.~Fernandez, S.~Yazdani, Z.~Yin,
  Z.~Wang, X.~Wang, W.~Xu, J.~Liu, et~al.
\newblock Optimization and improvement of fake news detection using voting
  technique for societal benefit.
\newblock In \emph{2023 IEEE International Conference on Data Mining Workshops
  (ICDMW)}, pages 1565--1574. IEEE, 2023.

\bibitem[Chinta et~al.(2024)Chinta, Wang, Yin, Hoang, Gonzalez, Quy, and
  Zhang]{chinta2024fairaied}
S.~V. Chinta, Z.~Wang, Z.~Yin, N.~Hoang, M.~Gonzalez, T.~L. Quy, and W.~Zhang.
\newblock Fairaied: Navigating fairness, bias, and ethics in educational ai
  applications.
\newblock \emph{arXiv preprint arXiv:2407.18745}, 2024.

\bibitem[Chinta et~al.(2025)Chinta, Wang, Palikhe, Zhang, Kashif, Smith, Liu,
  and Zhang]{chinta2025ai}
S.~V. Chinta, Z.~Wang, A.~Palikhe, X.~Zhang, A.~Kashif, M.~A. Smith, J.~Liu,
  and W.~Zhang.
\newblock Ai-driven healthcare: Fairness in ai healthcare: A survey.
\newblock \emph{PLOS Digital Health}, 4\penalty0 (5):\penalty0 e0000864, 2025.

\bibitem[Chu et~al.(2024{\natexlab{a}})Chu, Ni, Wang, Feng, Li, Hu, Xu, Yang,
  and Zhang]{chu2024history}
Z.~Chu, S.~Ni, Z.~Wang, X.~Feng, C.~Li, X.~Hu, R.~Xu, M.~Yang, and W.~Zhang.
\newblock History, development, and principles of large language models-an
  introductory survey.
\newblock \emph{arXiv preprint arXiv:2402.06853}, 2024{\natexlab{a}}.

\bibitem[Chu et~al.(2024{\natexlab{b}})Chu, Wang, and Zhang]{chu2024fairness}
Z.~Chu, Z.~Wang, and W.~Zhang.
\newblock Fairness in large language models: A taxonomic survey.
\newblock \emph{ACM SIGKDD Explorations Newsletter, 2024}, pages 34--48,
  2024{\natexlab{b}}.

\bibitem[Doan et~al.(2024)Doan, Wang, Hoang, and Zhang]{doan2024fairness}
T.~V. Doan, Z.~Wang, N.~N.~M. Hoang, and W.~Zhang.
\newblock Fairness in large language models in three hours.
\newblock In \emph{Proceedings of the 33rd ACM International Conference on
  Information and Knowledge Management}, pages 5514--5517, 2024.

\bibitem[Dzuong et~al.(2024)Dzuong, Wang, and Zhang]{dzuong2024uncertain}
J.~Dzuong, Z.~Wang, and W.~Zhang.
\newblock Uncertain boundaries: Multidisciplinary approaches to copyright
  issues in generative ai.
\newblock \emph{arXiv preprint arXiv:2404.08221}, 2024.

\bibitem[Guyet et~al.(2022)Guyet, Zhang, and Bifet]{guyet2022incremental}
T.~Guyet, W.~Zhang, and A.~Bifet.
\newblock Incremental mining of frequent serial episodes considering multiple
  occurrences.
\newblock In \emph{22nd International Conference on Computational Science},
  pages 460--472. Springer, 2022.

\bibitem[Hartwig et~al.(2024)Hartwig, Engel, Sick, Kniesel, Payer, Poonam,
  Gl{\"o}ckler, B{\"a}uerle, and Ropinski]{hartwig2024survey}
S.~Hartwig, D.~Engel, L.~Sick, H.~Kniesel, T.~Payer, P.~Poonam,
  M.~Gl{\"o}ckler, A.~B{\"a}uerle, and T.~Ropinski.
\newblock A survey on quality metrics for text-to-image generation.
\newblock \emph{arXiv preprint arXiv:2403.11821}, 2024.

\bibitem[Hayes(2023)]{hayes2023generative}
C.~M. Hayes.
\newblock Generative artificial intelligence and copyright: Both sides of the
  black box.
\newblock \emph{Available at SSRN 4517799}, 2023.

\bibitem[Joyce et~al.(2016)Joyce, Ochoa, Carroll, Leaffer, and
  Jaszi]{joyce2016copyright}
C.~Joyce, T.~T. Ochoa, M.~W. Carroll, M.~A. Leaffer, and P.~Jaszi.
\newblock \emph{Copyright law}, volume~85.
\newblock Carolina Academic Press Durham, NC, 2016.

\bibitem[Kim et~al.(2024)Kim, Min, Patel, Cheng, and Yang]{kim2024wouaf}
C.~Kim, K.~Min, M.~Patel, S.~Cheng, and Y.~Yang.
\newblock Wouaf: Weight modulation for user attribution and fingerprinting in
  text-to-image diffusion models.
\newblock In \emph{Proceedings of the IEEE/CVF Conference on Computer Vision
  and Pattern Recognition}, pages 8974--8983, 2024.

\bibitem[Le()]{le2024copyright}
T.~P.~V. Le.
\newblock Copyright protection towards generative ai artworks: The “clash”
  between us v. china and the implications for the european union.

\bibitem[Lee et~al.(2024)Lee, Adamopoulos, Todri, and Ghose]{lee2024impact}
H.~Lee, P.~Adamopoulos, V.~G. Todri, and A.~Ghose.
\newblock The impact of generative ai on advertising effectiveness.
\newblock 2024.

\bibitem[Liao(2025)]{liao2025deepseek}
H.~Liao.
\newblock Deepseek large-scale model: technical analysis and development
  prospect.
\newblock \emph{Journal of Computer Science and Electrical Engineering},
  7\penalty0 (1):\penalty0 33--37, 2025.

\bibitem[Liu et~al.(2021)Liu, Wang, Japkowicz, Tang, Zhang, and
  Zhao]{liu2021research}
Z.~Liu, R.~Wang, N.~Japkowicz, D.~Tang, W.~Zhang, and J.~Zhao.
\newblock Research on unsupervised feature learning for android malware
  detection based on restricted boltzmann machines.
\newblock \emph{Future Generation Computer Systems}, 120:\penalty0 91--108,
  2021.

\bibitem[Liu et~al.(2023)Liu, Wang, Japkowicz, Gomes, Peng, and
  Zhang]{liu2023segdroid}
Z.~Liu, R.~Wang, N.~Japkowicz, H.~M. Gomes, B.~Peng, and W.~Zhang.
\newblock Segdroid: An android malware detection method based on sensitive
  function call graph learning.
\newblock \emph{Expert Systems with Applications}, page 121125, 2023.

\bibitem[Lucchi(2024)]{lucchi2024chatgpt}
N.~Lucchi.
\newblock Chatgpt: a case study on copyright challenges for generative
  artificial intelligence systems.
\newblock \emph{European Journal of Risk Regulation}, 15\penalty0 (3):\penalty0
  602--624, 2024.

\bibitem[Pallant(2024)]{pallant2024creating}
J.~Pallant.
\newblock \emph{Creating Images Using AI: A Step-by-Step Guide to Midjourney}.
\newblock CRC Press, 2024.

\bibitem[Pizzi et~al.(2022)Pizzi, Roy, Ravindra, Goyal, and
  Douze]{pizzi2022self}
E.~Pizzi, S.~D. Roy, S.~N. Ravindra, P.~Goyal, and M.~Douze.
\newblock A self-supervised descriptor for image copy detection.
\newblock In \emph{Proceedings of the IEEE/CVF Conference on Computer Vision
  and Pattern Recognition}, pages 14532--14542, 2022.

\bibitem[Poland(2023)]{poland2023generative}
C.~M. Poland.
\newblock Generative ai and us intellectual property law.
\newblock \emph{arXiv preprint arXiv:2311.16023}, 2023.

\bibitem[Radford et~al.(2021)Radford, Kim, Hallacy, Ramesh, Goh, Agarwal,
  Sastry, Askell, Mishkin, Clark, et~al.]{radford2021learning}
A.~Radford, J.~W. Kim, C.~Hallacy, A.~Ramesh, G.~Goh, S.~Agarwal, G.~Sastry,
  A.~Askell, P.~Mishkin, J.~Clark, et~al.
\newblock Learning transferable visual models from natural language
  supervision.
\newblock In \emph{International conference on machine learning}, pages
  8748--8763. PmLR, 2021.

\bibitem[Rao et~al.(2024)Rao, Xiangli, Guo, Tang, Meng, and
  Agrawala]{rao2024generative}
A.~Rao, Y.~Xiangli, Y.~Guo, M.~Tang, C.~Meng, and M.~Agrawala.
\newblock Generative models for visual content editing and creation.
\newblock In \emph{ACM SIGGRAPH 2024 Courses}, pages 1--6. 2024.

\bibitem[Saharia et~al.(2022)Saharia, Chan, Chang, Lee, Ho, Salimans, Fleet,
  and Norouzi]{saharia2022palette}
C.~Saharia, W.~Chan, H.~Chang, C.~Lee, J.~Ho, T.~Salimans, D.~Fleet, and
  M.~Norouzi.
\newblock Palette: Image-to-image diffusion models.
\newblock In \emph{ACM SIGGRAPH 2022 conference proceedings}, pages 1--10,
  2022.

\bibitem[Samuelson(2023)]{samuelson2023ongoing}
P.~Samuelson.
\newblock Ongoing lawsuits could affect everyone who uses generative ai.
\newblock \emph{Science}, 381:\penalty0 6654, 2023.

\bibitem[Sandiumenge(2023)]{sandiumenge2023copyright}
I.~Sandiumenge.
\newblock Copyright implications of the use of generative ai.
\newblock \emph{Available at SSRN 4531912}, 2023.

\bibitem[Saxena et~al.(2023)Saxena, Zhang, and Shahabi]{saxena2023missed}
N.~A. Saxena, W.~Zhang, and C.~Shahabi.
\newblock Missed opportunities in fair ai.
\newblock In \emph{Proceedings of the 2023 SIAM International Conference on
  Data Mining (SDM)}, pages 961--964. SIAM, 2023.

\bibitem[Schuhmann et~al.(2022)Schuhmann, Beaumont, Vencu, Gordon, Wightman,
  Cherti, Coombes, Katta, Mullis, Wortsman, et~al.]{schuhmann2022laion}
C.~Schuhmann, R.~Beaumont, R.~Vencu, C.~Gordon, R.~Wightman, M.~Cherti,
  T.~Coombes, A.~Katta, C.~Mullis, M.~Wortsman, et~al.
\newblock Laion-5b: An open large-scale dataset for training next generation
  image-text models.
\newblock \emph{Advances in neural information processing systems}, 35, 2022.

\bibitem[Somepalli et~al.(2023{\natexlab{a}})Somepalli, Singla, Goldblum,
  Geiping, and Goldstein]{somepalli2023diffusion}
G.~Somepalli, V.~Singla, M.~Goldblum, J.~Geiping, and T.~Goldstein.
\newblock Diffusion art or digital forgery? investigating data replication in
  diffusion models.
\newblock In \emph{Proceedings of the IEEE/CVF conference on computer vision
  and pattern recognition}, pages 6048--6058, 2023{\natexlab{a}}.

\bibitem[Somepalli et~al.(2023{\natexlab{b}})Somepalli, Singla, Goldblum,
  Geiping, and Goldstein]{somepalli2023understanding}
G.~Somepalli, V.~Singla, M.~Goldblum, J.~Geiping, and T.~Goldstein.
\newblock Understanding and mitigating copying in diffusion models.
\newblock \emph{Advances in Neural Information Processing Systems},
  36:\penalty0 47783--47803, 2023{\natexlab{b}}.

\bibitem[Tang et~al.(2021)Tang, Zhang, Yu, Turner, Derr, Wang, and
  Ntoutsi]{tang2021interpretable}
X.~Tang, W.~Zhang, Y.~Yu, K.~Turner, T.~Derr, M.~Wang, and E.~Ntoutsi.
\newblock Interpretable visual understanding with cognitive attention network.
\newblock In \emph{International Conference on Artificial Neural Networks},
  pages 555--568. Springer, 2021.

\bibitem[Thongmeensuk(2024)]{thongmeensuk2024rethinking}
S.~Thongmeensuk.
\newblock Rethinking copyright exceptions in the era of generative ai:
  Balancing innovation and intellectual property protection.
\newblock \emph{The Journal of World Intellectual Property}, 27\penalty0
  (2):\penalty0 278--295, 2024.

\bibitem[Tirumala et~al.()Tirumala, Markosyan, Zettlemoyer, and
  Aghajanyan]{tirumala2022memorization}
K.~Tirumala, A.~Markosyan, L.~Zettlemoyer, and A.~Aghajanyan.
\newblock Memorization without overfitting: Analyzing the training dynamics of
  large language models.
\newblock volume~35.

\bibitem[Vartiainen and Tedre(2023)]{vartiainen2023using}
H.~Vartiainen and M.~Tedre.
\newblock Using artificial intelligence in craft education: crafting with
  text-to-image generative models.
\newblock \emph{Digital Creativity}, 34\penalty0 (1):\penalty0 1--21, 2023.

\bibitem[Vyas et~al.(2023)Vyas, Kakade, and Barak]{vyas2023provable}
N.~Vyas, S.~M. Kakade, and B.~Barak.
\newblock On provable copyright protection for generative models.
\newblock In \emph{International conference on machine learning}, pages
  35277--35299. PMLR, 2023.

\bibitem[Wang et~al.(2023{\natexlab{a}})Wang, Efros, Zhu, and
  Zhang]{wang2023evaluating}
S.-Y. Wang, A.~A. Efros, J.-Y. Zhu, and R.~Zhang.
\newblock Evaluating data attribution for text-to-image models.
\newblock In \emph{Proceedings of the IEEE/CVF International Conference on
  Computer Vision}, pages 7192--7203, 2023{\natexlab{a}}.

\bibitem[Wang et~al.()Wang, Zhang, Sun, and Yang]{wang2021bag}
W.~Wang, W.~Zhang, Y.~Sun, and Y.~Yang.
\newblock Bag of tricks and a strong baseline for image copy detection.
\newblock \emph{arXiv preprint arXiv:2111.08004}.

\bibitem[Wang et~al.(2024{\natexlab{a}})Wang, Sun, Tan, and
  Yang]{wang2024image}
W.~Wang, Y.~Sun, Z.~Tan, and Y.~Yang.
\newblock Image copy detection for diffusion models.
\newblock \emph{Advances in Neural Information Processing Systems},
  37:\penalty0 14417--14456, 2024{\natexlab{a}}.

\bibitem[Wang and Zhang(2024)]{wang2024group}
Z.~Wang and W.~Zhang.
\newblock Group fairness with individual and censorship constraints.
\newblock In \emph{27th European Conference on Artificial Intelligence}, 2024.

\bibitem[Wang and Zhang(2025)]{wang2025fdgen}
Z.~Wang and W.~Zhang.
\newblock Fdgen: A fairness-aware graph generation model.
\newblock In \emph{Proceedings of the 42nd International Conference on Machine
  Learning}. PMLR, 2025.

\bibitem[Wang et~al.(2023{\natexlab{b}})Wang, Narasimhan, Yao, and
  Zhang]{wang2023mitigating}
Z.~Wang, G.~Narasimhan, X.~Yao, and W.~Zhang.
\newblock Mitigating multisource biases in graph neural networks via real
  counterfactual samples.
\newblock In \emph{2023 IEEE International Conference on Data Mining (ICDM)},
  pages 638--647. IEEE, 2023{\natexlab{b}}.

\bibitem[Wang et~al.(2023{\natexlab{c}})Wang, Saxena, Yu, Karki, Zetty, Haque,
  Zhou, Kc, Stockwell, Bifet, et~al.]{wang2023preventing}
Z.~Wang, N.~Saxena, T.~Yu, S.~Karki, T.~Zetty, I.~Haque, S.~Zhou, D.~Kc,
  I.~Stockwell, A.~Bifet, et~al.
\newblock Preventing discriminatory decision-making in evolving data streams.
\newblock In \emph{Proceedings of the 2023 ACM Conference on Fairness,
  Accountability, and Transparency (FAccT)}, 2023{\natexlab{c}}.

\bibitem[Wang et~al.(2023{\natexlab{d}})Wang, Wallace, Bifet, Yao, and
  Zhang]{wang2023fg2an}
Z.~Wang, C.~Wallace, A.~Bifet, X.~Yao, and W.~Zhang.
\newblock Fg$^2$an: Fairness-aware graph generative adversarial networks.
\newblock In \emph{Joint European Conference on Machine Learning and Knowledge
  Discovery in Databases}, pages 259--275. Springer Nature Switzerland,
  2023{\natexlab{d}}.

\bibitem[Wang et~al.(2023{\natexlab{e}})Wang, Zhou, Haque, Lo, and
  Zhang]{wang2023towards}
Z.~Wang, Y.~Zhou, I.~Haque, D.~Lo, and W.~Zhang.
\newblock Towards fair machine learning software: Understanding and addressing
  model bias through counterfactual thinking.
\newblock \emph{arXiv preprint arXiv:2302.08018}, 2023{\natexlab{e}}.

\bibitem[Wang et~al.(2024{\natexlab{b}})Wang, Chen, Sehwag, Pan, and
  Lyu]{wang2024evaluating}
Z.~Wang, C.~Chen, V.~Sehwag, M.~Pan, and L.~Lyu.
\newblock Evaluating and mitigating ip infringement in visual generative ai.
\newblock \emph{arXiv preprint arXiv:2406.04662}, 2024{\natexlab{b}}.

\bibitem[Wang et~al.(2024{\natexlab{c}})Wang, Chu, Blanco, Chen, Chen, and
  Zhang]{wang2024advancing}
Z.~Wang, Z.~Chu, R.~Blanco, Z.~Chen, S.-C. Chen, and W.~Zhang.
\newblock Advancing graph counterfactual fairness through fair representation
  learning.
\newblock In \emph{Joint European Conference on Machine Learning and Knowledge
  Discovery in Databases}, pages 40--58. Springer Nature Switzerland,
  2024{\natexlab{c}}.

\bibitem[Wang et~al.(2024{\natexlab{d}})Wang, Chu, Doan, Ni, Yang, and
  Zhang]{wang2024history}
Z.~Wang, Z.~Chu, T.~V. Doan, S.~Ni, M.~Yang, and W.~Zhang.
\newblock History, development, and principles of large language models-an
  introductory survey.
\newblock \emph{AI and Ethics, 2024}, 2024{\natexlab{d}}.

\bibitem[Wang et~al.(2024{\natexlab{e}})Wang, Dzuong, Yuan, Chen, Wu, Yao, and
  Zhang]{wang2024individual1}
Z.~Wang, J.~Dzuong, X.~Yuan, Z.~Chen, Y.~Wu, X.~Yao, and W.~Zhang.
\newblock Individual fairness with group awareness under uncertainty.
\newblock In \emph{Joint European Conference on Machine Learning and Knowledge
  Discovery in Databases}, pages 89--106. Springer Nature Switzerland,
  2024{\natexlab{e}}.

\bibitem[Wang et~al.(2024{\natexlab{f}})Wang, Palikhe, Yin, and
  Zhang]{doan2024fairness1}
Z.~Wang, A.~Palikhe, Z.~Yin, and W.~Zhang.
\newblock Fairness definitions in language models explained.
\newblock \emph{arXiv preprint arXiv:2407.18454}, 2024{\natexlab{f}}.

\bibitem[Wang et~al.(2024{\natexlab{g}})Wang, Qiu, Chen, Salem, Yao, and
  Zhang]{wang2024toward}
Z.~Wang, M.~Qiu, M.~Chen, M.~B. Salem, X.~Yao, and W.~Zhang.
\newblock Toward fair graph neural networks via real counterfactual samples.
\newblock \emph{Knowledge and Information Systems}, pages 1--25,
  2024{\natexlab{g}}.

\bibitem[Wang et~al.(2024{\natexlab{h}})Wang, Ulloa, Yu, Rangaswami, Yap, and
  Zhang]{wang2024individual}
Z.~Wang, D.~Ulloa, T.~Yu, R.~Rangaswami, R.~Yap, and W.~Zhang.
\newblock Individual fairness with group constraints in graph neural networks.
\newblock In \emph{27th European Conference on Artificial Intelligence},
  2024{\natexlab{h}}.

\bibitem[Wang et~al.(2025{\natexlab{a}})Wang, Chu, Viet~Doan, Wang, Wu, Palade,
  and Zhang]{wang2025fair}
Z.~Wang, Z.~Chu, T.~Viet~Doan, S.~Wang, Y.~Wu, V.~Palade, and W.~Zhang.
\newblock Fair graph u-net: A fair graph learning framework integrating group
  and individual awareness.
\newblock In \emph{proceedings of the AAAI conference on artificial
  intelligence}, volume~39, pages 28485--28493, 2025{\natexlab{a}}.

\bibitem[Wang et~al.(2025{\natexlab{b}})Wang, Hoang, Zhang, Bello, Zhang,
  Iyengar, and Zhang]{wang2024towards}
Z.~Wang, N.~Hoang, X.~Zhang, K.~Bello, X.~Zhang, S.~S. Iyengar, and W.~Zhang.
\newblock Towards fair graph learning without demographic information.
\newblock In \emph{The 28th International Conference on Artificial Intelligence
  and Statistics}, volume 258, pages 2107--2115, 2025{\natexlab{b}}.

\bibitem[Wang et~al.(2025{\natexlab{c}})Wang, Liu, Pan, Liu, Saeed, Qiu, and
  Zhang]{wang2025fairgnn}
Z.~Wang, F.~Liu, S.~Pan, J.~Liu, F.~Saeed, M.~Qiu, and W.~Zhang.
\newblock fairgnn-wod: Fair graph learning without complete demographics.
\newblock In \emph{Proceedings of the 34th International Joint Conference on
  Artificial Intelligence}, 2025{\natexlab{c}}.

\bibitem[Wang et~al.(2025{\natexlab{d}})Wang, Wu, Moniz, Hu, Knijnenburg, Zhu,
  and Zhang]{wang2025limiteddemographics}
Z.~Wang, A.~Wu, N.~Moniz, S.~Hu, B.~Knijnenburg, Q.~Zhu, and W.~Zhang.
\newblock Towards fairness with limited demographics via disentangled learning.
\newblock In \emph{Proceedings of the 34th International Joint Conference on
  Artificial Intelligence}, 2025{\natexlab{d}}.

\bibitem[Wang et~al.(2025{\natexlab{e}})Wang, Yin, Liu, Liu, Lisetti, Yu, Wang,
  Liu, Ganapati, Zhou, et~al.]{wang2025graph}
Z.~Wang, Z.~Yin, F.~Liu, Z.~Liu, C.~Lisetti, R.~Yu, S.~Wang, J.~Liu,
  S.~Ganapati, S.~Zhou, et~al.
\newblock Graph fairness via authentic counterfactuals: Tackling structural and
  causal challenges.
\newblock \emph{ACM SIGKDD Explorations Newsletter}, 26\penalty0 (2):\penalty0
  89--98, 2025{\natexlab{e}}.

\bibitem[Wang et~al.(2025{\natexlab{f}})Wang, Yin, Palikhe, and
  Zhang]{wang2025FairnessT}
Z.~Wang, Z.~Yin, A.~Palikhe, and W.~Zhang.
\newblock Fairness in language models: A tutorial.
\newblock In \emph{Proceedings of the 34th ACM International Conference on
  Information and Knowledge Management}, 2025{\natexlab{f}}.

\bibitem[Wang et~al.(2025{\natexlab{g}})Wang, Yin, Yang, Zhuang, Yu, Kong, and
  Zhang]{wang2025Fairness}
Z.~Wang, Z.~Yin, L.~Yang, J.~Zhuang, R.~Yu, Q.~Kong, and W.~Zhang.
\newblock Fairness-aware graph representation learning with limited demographic
  information.
\newblock In \emph{Joint European Conference on Machine Learning and Knowledge
  Discovery in Databases}. Springer Nature Switzerland, 2025{\natexlab{g}}.

\bibitem[Wang et~al.(2025{\natexlab{h}})Wang, Yin, Yap, and Zhang]{wang2025AI}
Z.~Wang, Z.~Yin, R.~Yap, and W.~Zhang.
\newblock Ai fairness beyond complete demographics: Current achievements and
  future directions.
\newblock In \emph{28th European Conference on Artificial Intelligence},
  2025{\natexlab{h}}.

\bibitem[Wang et~al.(2025{\natexlab{i}})Wang, Yin, Yap, Zhang, Hu, and
  Zhang]{wang2025Redefining}
Z.~Wang, Z.~Yin, R.~Yap, X.~Zhang, S.~Hu, and W.~Zhang.
\newblock Redefining fairness: A multi-dimensional perspective and integrated
  evaluation framework.
\newblock In \emph{Joint European Conference on Machine Learning and Knowledge
  Discovery in Databases}. Springer Nature Switzerland, 2025{\natexlab{i}}.

\bibitem[Wang et~al.(2025{\natexlab{j}})Wang, Yin, Zhang, Zhang, He, Wang,
  Song, Yang, and Zhang]{wang2025towards}
Z.~Wang, Z.~Yin, X.~Zhang, Y.~Zhang, X.~He, S.~Wang, H.~Song, L.~Yang, and
  W.~Zhang.
\newblock Towards fair graph-based machine learning software: unveiling and
  mitigating graph model bias.
\newblock \emph{AI and Ethics}, pages 1--18, 2025{\natexlab{j}}.

\bibitem[Wang et~al.(2025{\natexlab{k}})Wang, Yin, Zhang, Yang, Zhang,
  Pissinou, Cai, Hu, Li, Zhao, et~al.]{wang2025fg}
Z.~Wang, Z.~Yin, Y.~Zhang, L.~Yang, T.~Zhang, N.~Pissinou, Y.~Cai, S.~Hu,
  Y.~Li, L.~Zhao, et~al.
\newblock Fg-smote: Towards fair node classification with graph neural network.
\newblock \emph{ACM SIGKDD Explorations Newsletter}, 26\penalty0 (2):\penalty0
  99--108, 2025{\natexlab{k}}.

\bibitem[Xu et~al.(2025)Xu, Wang, He, Han, and Tang]{xu2025can}
Q.~Xu, Z.~Wang, X.~He, L.~Han, and R.~Tang.
\newblock Can large vision-language models detect images copyright infringement
  from genai?
\newblock \emph{arXiv preprint arXiv:2502.16618}, 2025.

\bibitem[Yazdani et~al.(2024)Yazdani, Saxena, Wang, Wu, and
  Zhang]{yazdani2024comprehensive}
S.~Yazdani, N.~Saxena, Z.~Wang, Y.~Wu, and W.~Zhang.
\newblock A comprehensive survey of image and video generative ai: recent
  advances, variants, and applications.
\newblock 2024.

\bibitem[Yin et~al.(2024{\natexlab{a}})Yin, Agarwal, Kashif, Gonzalez, Wang,
  Liu, Liu, Wu, Stockwell, Xu, et~al.]{yin2024accessible}
Z.~Yin, S.~Agarwal, A.~Kashif, M.~Gonzalez, Z.~Wang, S.~Liu, Z.~Liu, Y.~Wu,
  I.~Stockwell, W.~Xu, et~al.
\newblock Accessible health screening using body fat estimation by image
  segmentation.
\newblock In \emph{2024 IEEE International Conference on Data Mining Workshops
  (ICDMW)}, pages 405--414, 2024{\natexlab{a}}.

\bibitem[Yin et~al.(2024{\natexlab{b}})Yin, Wang, and Zhang]{yin2024improving}
Z.~Yin, Z.~Wang, and W.~Zhang.
\newblock Improving fairness in machine learning software via counterfactual
  fairness thinking.
\newblock In \emph{Proceedings of the 2024 IEEE/ACM 46th International
  Conference on Software Engineering: Companion Proceedings}, pages 420--421,
  2024{\natexlab{b}}.

\bibitem[Yin et~al.(2025{\natexlab{a}})Yin, Wang, Palikhe, and
  Zhang]{yin2025Uncertain}
Z.~Yin, Z.~Wang, A.~Palikhe, and W.~Zhang.
\newblock Uncertain boundaries: A tutorial on copyright challenges and
  cross-disciplinary solutions for generative ai.
\newblock In \emph{Proceedings of the 34th ACM International Conference on
  Information and Knowledge Management}, 2025{\natexlab{a}}.

\bibitem[Yin et~al.(2025{\natexlab{b}})Yin, Wang, Xu, Zhuang, Mozumder, Smith,
  and Zhang]{yin2025digital}
Z.~Yin, Z.~Wang, W.~Xu, J.~Zhuang, P.~Mozumder, A.~Smith, and W.~Zhang.
\newblock Digital forensics in the age of large language models.
\newblock \emph{arXiv preprint arXiv:2504.02963}, 2025{\natexlab{b}}.

\bibitem[Zhang et~al.(2024)Zhang, Wang, Xu, Wang, and Shi]{zhang2024forget}
G.~Zhang, K.~Wang, X.~Xu, Z.~Wang, and H.~Shi.
\newblock Forget-me-not: Learning to forget in text-to-image diffusion models.
\newblock In \emph{Proceedings of the IEEE/CVF conference on computer vision
  and pattern recognition}, pages 1755--1764, 2024.

\bibitem[Zhang et~al.(2023{\natexlab{a}})Zhang, Nakamura, Isohara, and
  Sakurai]{zhang2023review}
H.~Zhang, T.~Nakamura, T.~Isohara, and K.~Sakurai.
\newblock A review on machine unlearning.
\newblock \emph{SN Computer Science}, 4\penalty0 (4):\penalty0 337,
  2023{\natexlab{a}}.

\bibitem[Zhang et~al.(2025{\natexlab{a}})Zhang, Wang, Palikhe, Yin, and
  Zhang]{zhang2025datasets}
J.~Zhang, Z.~Wang, A.~Palikhe, Z.~Yin, and W.~Zhang.
\newblock Datasets for fairness in language models: An in-depth survey.
\newblock \emph{arXiv preprint arXiv:2506.23411}, 2025{\natexlab{a}}.

\bibitem[Zhang et~al.(2025{\natexlab{b}})Zhang, Wang, Palikhe, Yin, and
  Zhang]{zhang2025datasetsfairnesslanguagemodels}
J.~Zhang, Z.~Wang, A.~Palikhe, Z.~Yin, and W.~Zhang.
\newblock Datasets for fairness in language models: An in-depth survey,
  2025{\natexlab{b}}.
\newblock URL \url{https://arxiv.org/abs/2506.23411}.

\bibitem[Zhang et~al.(2021{\natexlab{a}})Zhang, Zhang, Zhang, Chaddad, Guo,
  Zhang, Zhang, and Evans]{zhang2021autoencoder}
M.~Zhang, F.~Zhang, J.~Zhang, A.~Chaddad, F.~Guo, W.~Zhang, J.~Zhang, and
  A.~Evans.
\newblock Autoencoder for neuroimage.
\newblock In \emph{International conference on database and expert systems
  applications}, pages 84--90. Springer, 2021{\natexlab{a}}.

\bibitem[Zhang et~al.(2018{\natexlab{a}})Zhang, Isola, Efros, Shechtman, and
  Wang]{zhang2018unreasonable}
R.~Zhang, P.~Isola, A.~A. Efros, E.~Shechtman, and O.~Wang.
\newblock The unreasonable effectiveness of deep features as a perceptual
  metric.
\newblock In \emph{Proceedings of the IEEE conference on computer vision and
  pattern recognition}, pages 586--595, 2018{\natexlab{a}}.

\bibitem[Zhang(2020)]{zhang2020learning}
W.~Zhang.
\newblock Learning fairness and graph deep generation in dynamic environments.
\newblock 2020.

\bibitem[Zhang(2024{\natexlab{a}})]{zhang2024ai}
W.~Zhang.
\newblock Ai fairness in practice: Paradigm, challenges, and prospects.
\newblock \emph{Ai Magazine}, 45\penalty0 (3):\penalty0 386--395,
  2024{\natexlab{a}}.

\bibitem[Zhang(2024{\natexlab{b}})]{zhang2024fairness}
W.~Zhang.
\newblock Fairness with censorship: Bridging the gap between fairness research
  and real-world deployment.
\newblock In \emph{Proceedings of the AAAI Conference on Artificial
  Intelligence}, volume~38, pages 22685--22685, 2024{\natexlab{b}}.

\bibitem[Zhang(2025)]{zhang2025online}
W.~Zhang.
\newblock Online and customizable fairness-aware learning.
\newblock \emph{Knowledge and Information Systems}, pages 1--28, 2025.

\bibitem[Zhang and Ntoutsi(2019)]{zhang2019faht}
W.~Zhang and E.~Ntoutsi.
\newblock Faht: an adaptive fairness-aware decision tree classifier.
\newblock In \emph{Proceedings of the 28th International Joint Conference on
  Artificial Intelligence}, pages 1480--1486, 2019.

\bibitem[Zhang and Wang(2018)]{zhang2018content}
W.~Zhang and J.~Wang.
\newblock Content-bootstrapped collaborative filtering for medical article
  recommendations.
\newblock In \emph{IEEE International Conference on Bioinformatics and
  Biomedicine (BIBM)}, 2018.

\bibitem[Zhang and Weiss(2021)]{zhang2021fair}
W.~Zhang and J.~Weiss.
\newblock Fair decision-making under uncertainty.
\newblock In \emph{{2021 IEEE International Conference on Data Mining (ICDM)}}.
  IEEE, 2021.

\bibitem[Zhang and Weiss(2022)]{zhang2022longitudinal}
W.~Zhang and J.~C. Weiss.
\newblock Longitudinal fairness with censorship.
\newblock In \emph{proceedings of the AAAI conference on artificial
  intelligence}, volume~36, pages 12235--12243, 2022.

\bibitem[Zhang and Weiss(2023)]{zhang2023fairness}
W.~Zhang and J.~C. Weiss.
\newblock Fairness with censorship and group constraints.
\newblock \emph{Knowledge and Information Systems}, pages 1--24, 2023.

\bibitem[Zhang and Zhao(2020)]{zhang2020online}
W.~Zhang and L.~Zhao.
\newblock Online decision trees with fairness.
\newblock \emph{arXiv preprint arXiv:2010.08146}, 2020.

\bibitem[Zhang et~al.(2016)Zhang, Tang, and Wang]{zhang2016using}
W.~Zhang, J.~Tang, and N.~Wang.
\newblock Using the machine learning approach to predict patient survival from
  high-dimensional survival data.
\newblock In \emph{IEEE International Conference on Bioinformatics and
  Biomedicine (BIBM)}, 2016.

\bibitem[Zhang et~al.(2018{\natexlab{b}})Zhang, Wang, Jin, Oreopoulos, and
  Zhang]{zhang2018deterministic}
W.~Zhang, J.~Wang, D.~Jin, L.~Oreopoulos, and Z.~Zhang.
\newblock A deterministic self-organizing map approach and its application on
  satellite data based cloud type classification.
\newblock In \emph{IEEE International Conference on Big Data (Big Data)},
  2018{\natexlab{b}}.

\bibitem[Zhang et~al.(2019)Zhang, Tang, and Wang]{zhang2019fairness}
W.~Zhang, X.~Tang, and J.~Wang.
\newblock On fairness-aware learning for non-discriminative decision-making.
\newblock In \emph{International Conference on Data Mining Workshops (ICDMW)},
  pages 1072--1079, 2019.

\bibitem[Zhang et~al.(2021{\natexlab{b}})Zhang, Bifet, Zhang, Weiss, and
  Nejdl]{zhang2021farf}
W.~Zhang, A.~Bifet, X.~Zhang, J.~C. Weiss, and W.~Nejdl.
\newblock Farf: A fair and adaptive random forests classifier.
\newblock In \emph{Pacific-Asia Conference on Knowledge Discovery and Data
  Mining}, pages 245--256. Springer, 2021{\natexlab{b}}.

\bibitem[Zhang et~al.(2021{\natexlab{c}})Zhang, Zhang, Pfoser, and
  Zhao]{zhang2021disentangled}
W.~Zhang, L.~Zhang, D.~Pfoser, and L.~Zhao.
\newblock Disentangled dynamic graph deep generation.
\newblock In \emph{Proceedings of the SIAM International Conference on Data
  Mining (SDM)}, pages 738--746, 2021{\natexlab{c}}.

\bibitem[Zhang et~al.(2022)Zhang, Pan, Zhou, Walsh, and
  Weiss]{zhang2022fairness}
W.~Zhang, S.~Pan, S.~Zhou, T.~Walsh, and J.~C. Weiss.
\newblock Fairness amidst non-iid graph data: Current achievements and future
  directions.
\newblock \emph{arXiv preprint arXiv:2202.07170}, 2022.

\bibitem[Zhang et~al.(2023{\natexlab{b}})Zhang, Hernandez-Boussard, and
  Weiss]{zhang2023censored}
W.~Zhang, T.~Hernandez-Boussard, and J.~Weiss.
\newblock Censored fairness through awareness.
\newblock In \emph{Proceedings of the AAAI conference on artificial
  intelligence}, volume~37, pages 14611--14619, 2023{\natexlab{b}}.

\bibitem[Zhang et~al.(2023{\natexlab{c}})Zhang, Wang, Kim, Cheng, Oommen,
  Ravikumar, and Weiss]{zhang2023individual}
W.~Zhang, Z.~Wang, J.~Kim, C.~Cheng, T.~Oommen, P.~Ravikumar, and J.~Weiss.
\newblock Individual fairness under uncertainty.
\newblock In \emph{26th European Conference on Artificial Intelligence}, pages
  3042--3049, 2023{\natexlab{c}}.

\bibitem[Zhang et~al.(2025{\natexlab{c}})Zhang, Zhou, Walsh, and
  Weiss]{zhang2025fairness}
W.~Zhang, S.~Zhou, T.~Walsh, and J.~C. Weiss.
\newblock Fairness amidst non‐iid graph data: A literature review.
\newblock \emph{AI Magazine}, 46\penalty0 (1):\penalty0 e12212,
  2025{\natexlab{c}}.

\bibitem[Zhang et~al.(2020)]{zhang2020flexible}
W.~Zhang et~al.
\newblock Flexible and adaptive fairness-aware learning in non-stationary data
  streams.
\newblock In \emph{IEEE 32nd International Conference on Tools with Artificial
  Intelligence (ICTAI)}, pages 399--406, 2020.

\bibitem[Zhong et~al.(2023)Zhong, Chang, Yang, Wu, Mahawaga~Arachchige,
  Pathmabandu, and Xue]{zhong2023copyright}
H.~Zhong, J.~Chang, Z.~Yang, T.~Wu, P.~C. Mahawaga~Arachchige, C.~Pathmabandu,
  and M.~Xue.
\newblock Copyright protection and accountability of generative ai: Attack,
  watermarking and attribution.
\newblock In \emph{Companion Proceedings of the ACM Web Conference 2023}, pages
  94--98, 2023.

\end{thebibliography}

\end{document}